\documentclass[runningheads]{llncs}
\usepackage{graphicx}
\usepackage{amsmath,amssymb,gensymb}
\usepackage{color}
\usepackage[width=122mm,left=12mm,paperwidth=146mm,height=193mm,top=12mm,paperheight=217mm]{geometry}

\usepackage{multirow}
\usepackage{epsfig}
\usepackage{epstopdf}
\usepackage{bm}
\usepackage[font=small,labelfont=bf]{caption}
\usepackage{floatrow}
\newfloatcommand{capbtabbox}{table}[ ][\FBwidth]
\usepackage{subfig}
\usepackage{xspace}
\makeatletter
\DeclareRobustCommand\onedot{\futurelet\@let@token\@onedot}
\def\@onedot{\ifx\@let@token.\else.\null\fi\xspace}

\def\eg{\emph{e.g}\onedot} 
\def\ie{\emph{i.e}\onedot} 
 
\def\etc{\emph{etc}\onedot}

\makeatother

{}
{}
{}
{}

\newcommand{\cc}{}

\newcommand{\changes}{}

\newcommand{\KG}{\textcolor{black}}

\usepackage{wrapfig}
\usepackage{booktabs}

\begin{document}
\pagestyle{headings}
\mainmatter

\title{Look-ahead before you leap: end-to-end active recognition by forecasting the effect of motion}
\titlerunning{Look-ahead before you leap}
\authorrunning{Dinesh Jayaraman and Kristen Grauman}

\author{Dinesh Jayaraman \and Kristen Grauman}
\institute{
	The University of Texas at Austin\\
	\email{ \{dineshj,grauman\}@cs.utexas.edu}
}

\maketitle

\begin{abstract}
  Visual recognition systems mounted on autonomous moving agents face the challenge of unconstrained data, but simultaneously have the opportunity to improve their performance by moving to acquire new views of test data. In this work, we first show how a recurrent neural network-based system may be trained to perform end-to-end learning of motion policies suited for this ``active recognition'' setting. Further, we hypothesize that active vision requires an agent to have the capacity to reason about the effects of its motions on its view of the world. To verify this hypothesis, we attempt to induce this capacity in our active recognition pipeline, by simultaneously learning to forecast the effects of the agent's motions on \changes{its internal representation of the environment conditional on all past views}. Results across two challenging datasets confirm both that our end-to-end system successfully learns meaningful policies for active category recognition, and that ``learning to look ahead'' further boosts recognition performance.
\end{abstract}

\section{Introduction}

People consistently direct their senses in order to better understand their surroundings. For example, you might swivel around in your armchair to observe a person behind you, rotate a coffee mug on your desk to read an inscription, or walk to a window to observe the rain outside.

In sharp contrast to such scenarios, recent recognition research has been focused almost exclusively on static image recognition: the system takes a single snapshot as input, and produces a category label estimate as output.  The ease of collecting large labeled datasets of images has enabled major advances on this task in recent years, as evident for example in the striking gains made on the ImageNet challenge~\cite{ILSVRC15}.  Yet, despite this recent progress, recognition performance remains low for more complex, unconstrained images~\cite{lin2014microsoft}.

\begin{figure}[t]
  \centering
  \includegraphics[width=0.9\linewidth]{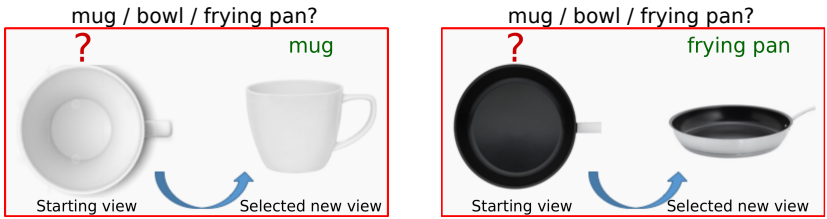}
  \caption{\small{A schematic illustrating the active categorization of two objects. A moving vision system may not recognize objects after just one view, but may intelligently choose to acquire new views to disambiguate amongst its competing hypotheses.}}
  \label{fig:conceptfig}
\end{figure}

Recognition systems mounted on autonomous moving agents acquire unconstrained visual input which may be difficult to recognize effectively, \emph{one frame at a time}.  However, similar to the human actor in the opening examples above, such systems have the opportunity to improve their performance by moving their camera apparatus or manipulating objects to acquire new information, as shown in Fig~\ref{fig:conceptfig}.  This control of the system over its sensory input has tremendous potential to improve its recognition performance.  While such mobile agent settings (mobile robots, autonomous vehicles, etc.) are closer to reality today than ever before, the problem of \emph{learning to actively move} to direct the acquisition of data remains underexplored in modern visual recognition research.

The problem we are describing fits into the realm of \emph{active vision}, which has a rich history in the literature (\emph{e.g.},~\cite{Andreopoulos2013-bm,wilkes1992active,Dickinson1997-my,Schiele1998-ph,callari2001active,aloimonos1988active}).  Active vision offers several technical challenges that are unaddressed in today's standard passive scenario.  In order to perform active vision, a system must learn to intelligently direct the acquisition of input to be processed by its recognition pipeline. In addition, recognition in an active setting places different demands on a system than in the standard passive scenario. To take one example, ``nuisance factors'' in still image recognition---such as pose, lighting, and viewpoint changes---become \emph{avoidable} factors in the active vision setting, since in principle, they can often be overcome merely by moving the agent to the right location.

This calls for a major change of approach.  Rather than strive for invariance to nuisance factors as is the standard in static image recognition, an intriguing  strategy is to learn to \emph{identify when conditions are non-ideal for recognition} and to \emph{actively select the correct agent motion} that will lead to better conditions. In addition, recognition decisions must be made based on intelligently fusing evidence from multiple observations.

We contend that these three functions of an active vision system---control, per-view recognition, and evidence fusion---are closely intertwined, and must be tailored to work together.  In particular, as the first contribution of this paper, we propose to learn all three modules of an active vision system simultaneously and end-to-end.  We employ a stochastic neural network to learn intelligent motion policies (control), a standard neural network to process inputs at each timestep (per-view recognition), and a modern recurrent neural network (RNN) to integrate evidence over time (evidence fusion).  Given an initial view and a set of possible agent motions, our approach learns how to move in the 3-D environment to produce accurate categorization results.

Additionally, we hypothesize that motion planning for active vision requires an agent to internally ``look before it leaps''.  That is, it ought to simultaneously reason about the effect of its motions on future inputs.  To demonstrate this, as a second contribution, we jointly train our active vision system to have the ability to predict \changes{\emph{how its internal representation of its environment will evolve} conditioned on its choice of motion}. As we will explain below, this may be seen as preferring equivariance \ie predictable feature responses to pose changes, rather than invariance as is standard in passive recognition pipelines.

Through experiments on two datasets, we validate both our key ideas: (1) RNN-based end-to-end active categorization and (2) learning to forecast the effects of self-motion at the same time one learns how to move to solve the recognition task.  We study both a scene categorization scenario, where the system chooses how to move around a previously unseen 3-D scene, and an object categorization scenario, where the system chooses how to manipulate a previously unseen object that it holds.  Our results establish the advantage of our end-to-end approach over both passive and traditional active methods.

\section{Related work}\label{sec:related}

\paragraph{Active vision} The idea that a subject's actions may play an important role in perception can be traced back almost 150 years~\cite{brentano1874psychologie} in the cognitive psychology literature~\cite{Andreopoulos2013-bm}. ``Active perception'', the idea of exploiting \emph{intelligent control strategies} (agent motion, object manipulation, camera parameter changes \etc) for \emph{goal-directed data acquisition} to improve machine vision, was pioneered by~\cite{bajcsy1988active,aloimonos1988active,ballard1991animate,wilkes1992active}. While most research in this area has targeted low-level vision problems such as segmentation, structure from motion, depth estimation, optical flow estimation~\cite{mishra2009active,ballard1991animate,aloimonos1988active}, or the ``semantic search'' task of object localization~\cite{Andreopoulos2009-mh,helmer2009semantic,gonzalez2014active,Soatto2009-pk}, approaches targeting \emph{active recognition} are most directly related to our work.

Most prior active recognition approaches attempt to identify during training those canonical/``special'' views that minimize ambiguity among candidate labels~\cite{wilkes1992active,Dickinson1997-my,Schiele1998-ph,callari2001active,Denzler2002-hf}.  At test time, such systems iteratively  estimate the current pose, then select moves that take them to such pre-identified informative viewpoints. These approaches are typically applicable only to \emph{instance} recognition problems, since broader categories can be too diverse in appearance and shape to fix ``special viewpoints''.

In contrast, our approach handles real world categories. To the best of our knowledge, very little prior work attempts this challenging task of active \emph{category} recognition (as opposed to instance recognition)~\cite{ramanathan2011active,wu20153d,Jayaraman2015-aq,aloim,borotschnig1998active}.
 The increased difficulty is due to the fact that with complex real world categories, it is much harder to anticipate new views conditioned on actions. Since new instances will be seen at test time, it is not sufficient to simply memorize the geometry of individual instances, as many active instance recognition methods effectively do.

 \changes{
   Recently,~\cite{wu20153d,Jayaraman2015-aq} learn to \emph{predict} the next views of \emph{unseen test objects}, and use this to explicitly greedily reason about the most informative ``next-best'' move. Instead, our approach uses reinforcement learning (RL) in a stochastic recurrent neural network to learn optimal \emph{sequential} movement policies over multiple timesteps. The closest methods to ours in this respect are~\cite{paletta2000active} and~\cite{GERMS}, both of which employ Q-learning in feedforward neural networks to perform view selection, and target relatively simpler visual tasks compared to this work.}

In addition to the above, an important novelty of our approach is in learning the entire system end-to-end.  Active recognition approaches must broadly perform three separate functions: action selection, per-instant view processing, and belief updates based on the history of observed views.  While previous approaches have explored several choices for action selection, they typically train a ``passive'' per-instant view recognition module offline and fuse predictions across time using some manually defined heuristic~\cite{Schiele1998-ph,Dickinson1997-my,Denzler2002-hf,ramanathan2011active,GERMS}.  For example, recently, a deep neural network is trained to learn action policies in~\cite{GERMS} after pretraining a per-view classifier and using a simple Naive Bayes update heuristic for label belief fusion. In contrast, we train all three modules jointly within a single active recognition objective. %

\paragraph{Saliency and attention}

Visual saliency and attention are related to active vision~\cite{Mnih2014-dg,Ba2014-rr,Xu2015-vc,Bazzani2011-cb,sermanet2014attention}. While active vision systems aim to form policies to acquire \emph{new} data, saliency and attention systems aim to block out ``distractors'' in \emph{existing} data by identifying portions of input images/video to focus on, \cc{often as a faster alternative to sliding window-based methods}. Attention systems thus sometimes take a ``foveated'' approach~\cite{butko,Mnih2014-dg}.  In contrast, in our setting, the system never holds a snapshot of the entire environment at once.  Rather, its input at each timestep is one portion of its complete physical 3D environment, and it must choose motions leading to more informative---possibly non-overlapping---viewpoints.  Another difference between the two settings is that the focus of attention may move in arbitrary jumps (saccades) without continuity, whereas active vision agents may only move continuously.

Sequential attention systems using recurrent neural networks in particular have seen significant interest of late~\cite{Mnih2014-dg}, with variants proving successful across several attention-based tasks~\cite{Ba2014-rr,Xu2015-vc,sermanet2014attention}. We adopt the basic attention architecture of~\cite{Mnih2014-dg} as a starting point for our model, and develop it further
to accommodate the active vision setting, instill look-ahead capabilities, and select camera motions surrounding a 3D object that will most facilitate categorization.

\paragraph{Predicting related features}

\changes{
There is recent interest in ``visual prediction'' problems in various contexts~\cite{Vondrick2015-vq,ranzato2014video,Kulkarni2015-uh,Ding2014-co,wu20153d,Flynn2015-fm,Jayaraman2015-aq,walker2015dense}, often using convolutional neural networks (CNNs).  For example, one can train CNNs~\cite{Vondrick2015-vq,ssfa} or recurrent neural networks (RNNs) to predict future frames based on previously observed frames~\cite{ranzato2014video} in an entirely passive setting.  These methods do not attempt to reason about \emph{causes} of  view transformations \eg camera motions. Closer to our work are methods for view synthesis, such as ~\cite{Kulkarni2015-uh,Ding2014-co}, which allow synthesis of simple synthetic images with specified ``factors of variation'' (such as pose and lighting).}  Given surrounding views, high quality unseen views are predicted in~\cite{Flynn2015-fm}, effectively learning 3D geometry end-to-end. The methods of~\cite{Jayaraman2015-aq,pulkit} model feature responses to a discrete set of observer motions.  Different from all the above, we learn to predict the evolution of temporally aggregated features---computed from a complete history of seen views---as a function of observer motion choices.  Furthermore, we integrate this idea with the closely tied active recognition problem.

\paragraph{Integrating sensors and actions}

\changes{Our work is also related to research in sensorimotor feature embeddings~\cite{Bowling2005-ev,Stober2011-sn,Jayaraman2015-aq,cohen2014transformation,chen2015deepdriving,Levine2015-gi,Watter2015-jn}.  There the idea is to combine (possibly non-visual) sensor streams together with proprioception or other knowledge about the actions of the agent on which the sensors are mounted.
Various methods learn features that  transform in simple ways in response to an agent's actions~\cite{Bowling2005-ev,Jayaraman2015-aq,cohen2014transformation} or reflect the geometry of an agent's environment~\cite{Stober2011-sn}. Neural nets are trained to perform simple robotic tasks in~\cite{chen2015deepdriving,Levine2015-gi}.
Perhaps conceptually most relevant to our work among these is~\cite{Watter2015-jn}.  Their method learns an image feature space to determine control actions easily from visual inputs, with applications to simulated control tasks.  In contrast, we learn embeddings encoding information from complete \emph{histories} of observations and agent actions, with the aim of exposing this information to an active visual recognition controller.
}

\section{Approach}\label{sec:approach}

First we define the setting and data flow for active recognition (Sec.~\ref{sec:setting}).  Then we define our basic system architecture (Sec.~\ref{sec:vanillaRAM}).  Finally, we describe our look-ahead module (Sec.~\ref{sec:lookahead_reg}).

\subsection{Setting}\label{sec:setting}

We first describe our active vision setting at test time, using a 3-D object category recognition scenario as a running example.   Our results consider both object and scene category recognition tasks.  The active recognition system can issue motor commands to move a camera within a viewing sphere around the 3-D object $X$ of interest. Each point on this viewing sphere is indexed by a corresponding 2-D camera pose vector $\bm{p}$ indexing elevation and azimuth.

The system is allowed $T$ timesteps to recognize every object instance $X$. At every timestep $t=1,2,\ldots T$:
\begin{itemize}
  \item The system issues a motor command $\bm{m}_t$ \emph{e.g.} ``increase camera elevation by 20\degree, azimuth by 10\degree'', from a set $\mathcal{M}$ of available camera motions. In our experiments, $\mathcal{M}$ is a discrete set consisting of small camera motions to points on an elevation-azimuth grid centered at the previous camera pose $\bm{p}_{t-1}$.
 At time $t=1$, the ``previous'' camera pose $p_0$ is set to some random unknown vector, corresponding to the agent initializing its recognition episode at some arbitrary position with respect to the object.
  \item Next, the system is presented a new 2-D view $\bm{x}_t=P(X,\bm{p}_t)$ of $X$ captured from the new camera pose $\bm{p}_t=\bm{p}_{t-1}+\bm{m}_t$, where $P(.,.)$ is a projection function. %
This new evidence is now available to the system while selecting its next action $\bm{m}_{t+1}$.
\end{itemize}
At the final timestep $t=T$, the system must additionally predict a category label $\hat{y}$ for $X$, \emph{e.g.}, the object category it believes is most probable.  In our implementation, the number of timesteps $T$ is fixed, and all valid motor commands have uniform cost.  The system is evaluated only on the accuracy of its prediction $\hat{y}$.  However, the framework generalizes to the case of variable-length episodes.

\subsection{Active recognition system  architecture}\label{sec:vanillaRAM}

\begin{figure}[t]
  \centering
  \includegraphics[width=0.6\linewidth]{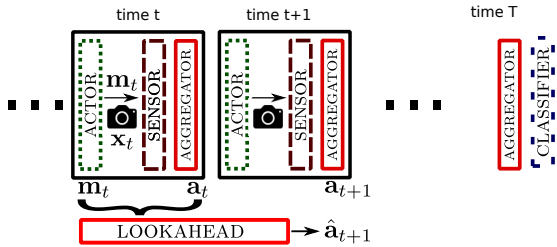}
  \caption{\small{A schematic of our system architecture depicting the interaction between \textsc{actor}, \textsc{sensor} and \textsc{aggregator} and \textsc{classifier} modules, unrolled over timesteps. Information flows from left to right. At training time, the additional \textsc{lookahead} acts across two timesteps, learning to predict the evolution of the aggregate feature $\bm{a}_t$ into $\bm{a}_{t+1}$ conditional on the selected motion $\bm{m}_t$. See Sec~\ref{sec:vanillaRAM} for details.}}
  \label{fig:flow}
\end{figure}

Our basic active recognition system is modeled on the recurrent architecture first proposed in~\cite{Mnih2014-dg} for visual attention.
Our system is composed of four basic modules: \textsc{actor}, \textsc{sensor}, \textsc{aggregator} and  \textsc{classifier}, with weights $W_a, W_s, W_r, W_c$ respectively. At each step $t$, \textsc{actor} issues a motor command $\bm{m}_t$, which updates the camera pose vector to $\bm{p}_t=\bm{p}_{t-1}+\bm{m}_t$. Next, a 2-D image $\bm{x}_t$ captured from this pose is fed into \textsc{sensor} together with the motor command $\bm{m}_t$. \textsc{sensor} produces a view-specific feature vector $\bm{s}_t=\textsc{sensor}(\bm{x}_t, \bm{m}_t)$, which is then fed into \textsc{aggregator} to produce aggregate feature vector $\bm{a}_t=\textsc{aggregator}(\bm{s}_1,\ldots,\bm{s}_t)$. The cycle is completed when, at the next step $t+1$, \textsc{actor} processes the aggregate feature from the previous timestep to issue $\bm{m}_{t+1}=\textsc{actor}(\bm{a}_t)$. Finally, after $T$ steps, the category label beliefs are predicted as $\hat{y}(W, X)=\textsc{classifier}(\bm{a}_t)$, where $W=[W_a, W_s, W_r, W_c]$ is the vector of all learnable weights in the network, and for a $C$-class classification problem, $\hat{y}$ is a $C$-dimensional multinomial probability density function representing the likelihoods of the 3-D object $X$ belonging to each of the $C$ classes. See Fig~\ref{fig:flow} for a schematic showing how the modules are connected. %

In our setup, \textsc{aggregator} is a recurrent neural network, \textsc{classifier} is a simple fully-connected hidden layer followed by a log-softmax and \textsc{sensor} separately processes the view $\bm{x}_t$ and the motor signal $\bm{m}_t$ in disjoint neural network pipelines before merging them through more layers of processing to produce the per-instance view feature $\bm{s}_t=\textsc{sensor}(\bm{x}_t, \bm{m}_t)$. \textsc{actor} has a non-standard neural net architecture involving stochastic units: at each timestep, it internally produces an $|\mathcal{M}|$-dimensional multinomial density function $\pi(\bm{m}_t)$ over all candidate camera motions in $\mathcal{M}$, from which it samples one motion. For more details on the internal architectures of these modules, see Supp.%

\paragraph{Training}
\changes{At training time, the network weights $W$ are trained jointly to maximize classifier accuracy at time $T$. Following~\cite{Mnih2014-dg}, training $W$ follows a hybrid procedure involving both standard backpropagation and ``connectionist reinforcement learning''~\cite{williams1992simple}. The modules with standard deterministic neural network connections (\textsc{classifier}, \textsc{aggregator} and \textsc{sensor}) can be trained directly by backpropagating gradients from a softmax classification loss, while the \textsc{actor} module which contains stochastic units can only be trained using the REINFORCE procedure of~\cite{williams1992simple}.
}

Roughly, REINFORCE treats the \textsc{actor} module as a Partially Observable Markov Decision Process (POMDP), with the pdf $\pi(\bm{m}_t|\bm{a}_{t-1}, W)$ representing the policy to be learned. In a reinforcement learning (RL)-style approach, REINFORCE iteratively increases weights in the pdf $\pi(\bm{m})$ on those candidate motions $\bm{m}\in \mathcal{M}$ that have produced higher ``rewards'', as defined by a reward function. A simple REINFORCE reward function to promote classification accuracy could be $R_c(\hat{y})=1$ when the most likely label in $\hat{y}$ is correct, and 0 when not. \cc{To speed up training, we use a variance-reduced version of this loss $R(\hat{y})=R_c(\hat{y})-R_c(z)$, where $z$ is set to the most commonly occuring class.}
Beyond the stochastic units, the REINFORCE algorithm produces gradients that may be propagated to non-stochastic units through standard backpropagation. In our hybrid training approach, these REINFORCE gradients from \textsc{actor} are therefore added to the softmax loss gradients from \textsc{classifier} before backpropagation through \textsc{aggregator} and \textsc{sensor}.

More formally, given a training dataset of instance-label pairs $\{(X^i, y^i): 1 \leq i \leq N\}$, the gradient updates are as follows. Let  $W_{\backslash c}$ denote $[W_a, W_s, W_r]$, \ie all the weights in $W$ except the \textsc{classifier} weights $W_c$, and similarly, let $W_{\backslash a}$ denote $[W_c, W_r, W_s]$. Then:
\begin{align}
  \Delta W_{\backslash c}^{RL} &\approx \sum_{i=1}^N \sum_{t=1}^T \nabla_{W_{\backslash c}}\log\pi(m_t^i | \bm{a}_{t-1}^i; W_{\backslash c}) R^i,
  \label{eq:vanilla_grad_update1}\\
  \Delta W_{\backslash a}^{SM} &= - \sum_{i=1}^N \nabla_{W_{\backslash a}} L_{\text{softmax}}(\hat{y}^i(W, X), y^i),
  \label{eq:vanilla_grad_update2}
\end{align}
where indices $i$ in the superscripts denote correspondence to the $i^{th}$ training sample $X^i$. Eq~\eqref{eq:vanilla_grad_update1} and~\eqref{eq:vanilla_grad_update2} show the gradients computed from the REINFORCE rewards (RL) and the softmax loss (SM) respectively, for different subsets of weights. The REINFORCE gradients $\Delta W^{RL}$ are computed using the approximation proposed in~\cite{williams1992simple}. Final gradients with respect to the weights of each module used in weight updates are given by: $\Delta W_a=\Delta W_a^{RL}$, $\Delta W_s = \Delta W_s^{RL}+\Delta W_s^{SM}$, $\Delta W_r= \Delta W_r ^{RL} + \Delta W_r^{SM}$, $\Delta W_c = \Delta W_c ^{RL} + \Delta W_c^{SM}$. Training is through standard stochastic gradient descent with early stopping based on a validation set.

\subsection{Look-ahead: predicting the effects of motions}\label{sec:lookahead_reg}
Active recognition systems select the next motion based on some expectation of the next view.
Though non-trivial even in the traditional instance recognition setting~\cite{wilkes1992active,Dickinson1997-my,Schiele1998-ph,callari2001active}, with instances one can exploit the fact that pose estimation in some canonical pose space is sufficient in itself to estimate properties of future views. In other words, with enough prior experience seeing the object instance, it is largely a 3-D (or implicit 3-D) geometric model formation problem.

In contrast, as discussed in Sec.~\ref{sec:related}, this problem is much harder in active \emph{categorization} with realistic categories---the domain we target.  Predicting subsequent views  in this setting is severely under-constrained, and requires reasoning about semantics and geometry together.  In other words, next view planning requires some element of learning about how 3-D objects \emph{in general} change in their appearance as a function of observer motion.

We hypothesize that the ability to predict the next view conditional on the next camera motion is closely tied to the ability to select optimal motions.   Thus, rather than learn separately the model of view transitions and model of motion policies, we propose a unified approach to learn them jointly.
Our idea is that knowledge transfer from a view prediction task will benefit active categorization.
In this formulation, we retain the system from Sec.~\ref{sec:vanillaRAM}, but simultaneously learn to predict, at every timestep $t$, the impact on aggregate features $\bm{a}_{t+1}$ at the next timestep, given $\bm{a}_t$ and any choice of motion $\bm{m}_t\in \mathcal{M}$.  In other words, we simultaneously learn how the \emph{accumulated history} of learned features---not only the current view---will evolve as a function of our candidate motions.

For this auxiliary task, we introduce an additional module, \textsc{lookahead}, with weights $W_l$ into the setup of Sec.~\ref{sec:vanillaRAM} at training time. At timestep $t$, \textsc{lookahead} takes as input $\bm{a}_{t-1}$ and $\bm{m}_{t-1}$ and predicts $\hat{\bm{a}}_t=\textsc{lookahead}(\bm{a}_{t-1}, \bm{m}_{t-1})$. This module may be thought of as a ``predictive auto-encoder'' in the space of aggregate features $\bm{a}_t$ output by \textsc{aggregator}. A look-ahead error loss is computed at every timestep between the predicted and actual aggregate features: $d(\hat{\bm{a}}_{t}, \bm{a}_t | \bm{a}_{t-1}, \bm{m}_{t-1})$. We use the cosine distance to compute this error. This per-timestep look-ahead loss provides a third source of training gradients $\Delta W_{\backslash ca}^{LA}$ for the network weights, as it is backpropagated through \textsc{aggregator} and \textsc{sensor}:%
\begin{equation}
  \Delta W_{\backslash ca}^{LA} = \sum_{i=1}^N \sum_{t=2}^T \nabla_{W_{\backslash ca}} d(\hat{\bm{a}}_{t}, \bm{a}_t | \bm{a}_{t-1}, \bm{m}_{t-1}),
  \label{eq:lookahead_grad_update}
\end{equation}
where $W$ now includes $W_l$ and $LA$ denotes lookahead. The \textsc{lookahead} module itself is trained solely from this error, so that $\Delta W_l= \Delta W_l^{LA}$. The final gradients used to train \textsc{sensor} and \textsc{aggregator} change to include this new loss: $\Delta W_s = \Delta W_s^{RL}+\Delta W_s^{SM}+\lambda\Delta W_s^{LA}$, $\Delta W_r= \Delta W_r ^{RL} + \Delta W_r^{SM}+ \lambda\Delta W_r^{LA}$. $\lambda$ is a new hyperparameter that controls how much the weights in the core network are influenced by the look-ahead error loss.

The look-ahead error loss of Eq~\ref{eq:lookahead_grad_update} may also be interpreted as an unsupervised regularizer on the classification objective of Eq~\ref{eq:vanilla_grad_update1} and \ref{eq:vanilla_grad_update2}. This regularizer encodes the hypothesis that good features for the active recognition task must respond in learnable, systematic ways to  camera motions.

This is related to the role of ``equivariant'' image features in~\cite{Jayaraman2015-aq}, where we showed that regularizing image features to respond predictably to observer egomotions improves performance on standard static image categorization tasks. This work differs from~\cite{Jayaraman2015-aq} in several important ways.  First, we explore the utility of look-ahead for the active categorization problem, not recognition of individual static images.  Second, the proposed look-ahead module is conceptually distinct.  In particular, we propose to regularize the aggregate features from a sequence of activity, not simply per-view features.  Whereas in~\cite{Jayaraman2015-aq} the effect of a discrete ego-motion on one image is estimated by linear transformations in the embedding space, the proposed look-ahead module takes as input both the history of views and the selected motion when estimating the effects of hypothetical motions.

\paragraph{Proprioceptive knowledge}
Another useful feature of our approach is that it allows for easy modeling of proprioceptive knowledge such as the current position $\bm{p}_t$ of a robotic arm. Since the \textsc{actor} module is trained purely through REINFORCE rewards, all other modules may access its output $\bm{m}_t$ without having to backpropagate extra gradients from the softmax loss. For instance, while the sensor module is fed $\bm{m}_t$ as input, it does not directly backpropagate any gradients to train \textsc{actor}. Since $\bm{p}_t$ is a function solely of $(\bm{m}_1 ... \bm{m}_t)$, this knowledge is readily available for use in other components of the system without any changes to the training procedure described above. We append appropriate proprioceptive information to the inputs of \textsc{actor} and \textsc{lookahead}, detailed in experiments.

\paragraph{Greedy softmax classification loss}
We found it beneficial at training time to inject softmax classification gradients after every timestep, rather than only at the end of $T$ timesteps. To achieve this, the $\textsc{classifier}$ module is modified to contain a bank of $T$ classification networks with identical architectures (but different weights, since in general, \textsc{aggregator} outputs $\bm{a}_t$ at different timesteps may have domain differences). Note that the REINFORCE loss is still computed only at $t=T$. Thus, given that softmax gradients do not pass through the \textsc{actor} module, it remains free to learn non-greedy motion policies.

\section{Experiments}\label{sec:exp}

We evaluate our approach for object and scene categorization.  In both cases, the system must choose how it will move in its 3-D environment such that the full sequence of its actions lead to the most accurate categorization results.

\subsection{Datasets and baselines}\label{sec:datasets}

While active vision systems have traditionally been tested on custom robotic setups~\cite{ramanathan2011active} (or simple turn-table-style datasets~\cite{Schiele1998-ph}), we aim to test our system on realistic, off-the-shelf datasets in the interest of benchmarking and reproducibility. We work with two publicly available datasets, SUN360~\cite{SUN360} and GERMS~\cite{GERMS}.

\begin{figure}[t]
  \centering
  \includegraphics[width=1\linewidth]{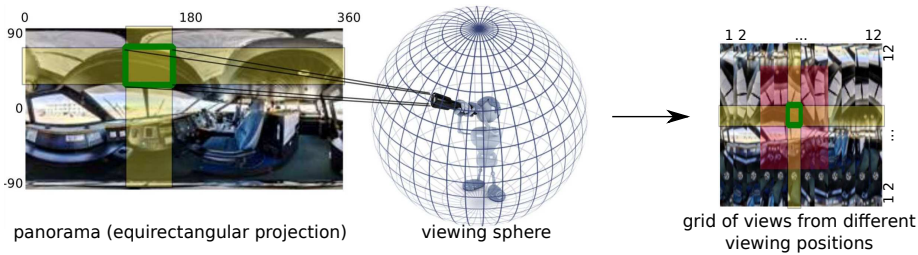}
  \caption{\small{(Best seen in color) An ``airplane interior'' class example showing how SUN360 spherical panoramas (equirectangular projection on the left) are converted into $12\times 12$ 45\degree FOV view grid.
As an illustration, the view at grid coordinates $x = 4$, $y = 6$ outlined in green
in the view grid on the right corresponds approximately to the overlap region (also outlined in green)
on the left (approximate because of panorama distortions---rectangles in the panorama are not rectangles in the rectified views present in the grid).
The $5\times 7$ red shaded region in the view grid (right) shows the motions available to \textsc{actor} when starting from the highlighted view.}}
  \label{fig:SUN360}
\end{figure}

Our \textbf{SUN360~\cite{SUN360}} experiments test a scenario where the agent is exploring a 3-D scene and must intelligently turn to see new parts of the scene that will enable accurate scene categorization (bedroom, living room, etc.).  SUN360 consists of spherical panoramas of various indoor and outdoor scenes together with scene category labels. We use the 26-category subset (8992 panoramic images) used in~\cite{SUN360}.  Each panorama by itself represents a 3-D scene instance, around which an agent ``moves'' by rotating its head, as shown in Fig~\ref{fig:SUN360}.  For our experiments, the agent has a limited field of view (45\degree) at each timestep.  We sample discrete views in a 12 elevations (camera pitch) $\times$ 12 azimuths (camera yaw) grid. The pitch and yaw steps are both spaced 30\degree apart (12$\times$30=360), so that the entire viewing sphere is uniformly sampled on each axis. Starting from a full panorama of size $1024\times 2048$, each 45\degree~FOV view is represented first as a $224\times 224$ image, from which $1024$-dim.~GoogleNet~\cite{googlenet} features are extracted from the penultimate layer. At each timestep, the agent can choose to move to viewpoints on a 5$\times$7 grid centered at the current position. We set $T=5$ timesteps.\footnote{Episode lengths were set based on learning time for efficient experimentation.} Proprioceptive knowledge in the form of the current camera elevation angle is fed into \textsc{actor} and \textsc{lookahead}. We use a random 80-20 train-test split.  Our use of SUN360 to simulate an active agent in a 3D scene is new and offers a realistic scenario that we can benchmark rigorously; note that previous work on the dataset does a different task, \emph{i.e.}, recognition with the full panorama in hand at once~\cite{SUN360}, and results are therefore not comparable to our setting.

\begin{figure}[t]
  \centering
  \includegraphics[width=1\linewidth]{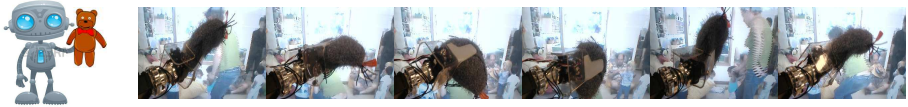}
  \caption{\small{The GERMS active object instance recognition dataset~\cite{GERMS} contains videos of a single-axis robotic hand rotating 136 toys against a moving background.}}
  \label{fig:GERMS}
\end{figure}

Our \textbf{GERMS~\cite{GERMS}} experiments consider the scenario where a robot is holding an object and must decide on its next best motion relative to that object, \emph{e.g.}, to gain access to an unseen facet of the object, so as to recognize its instance label.  %
GERMS has 6 videos each (3 train, 3 test) of 136 objects being rotated around different fixed axes, against a television screen displaying moving indoor scenes (see Fig~\ref{fig:GERMS}).  Each video frame is annotated by the angle at which the robotic arm is holding the object.  Each video provides one collection of views that our active vision system can traverse at will, for a total of $136\times 6=816$ train/test instances (compared to 8992 on SUN360). While GERMS is small and targets \emph{instance} rather than category recognition, aside from SUN360 it is the most suitable prior dataset facilitating active recognition. Each frame is represented by a 4096-dim.~VGG-net feature vector~\cite{simonyan2014very}, provided by the authors~\cite{GERMS}.   We set episode lengths to $T=3$ steps.  As proprioceptive knowledge, we feed the current position of the robotic hand into \textsc{actor} and \textsc{lookahead}.  We use the train-test subsets specified by the dataset authors.

\paragraph{Baselines:}

We extensively evaluate our ``Look-ahead active RNN'' (Sec~\ref{sec:lookahead_reg}) and simpler ``Active RNN'' (Sec~\ref{sec:vanillaRAM}) against six baselines, including passive single-view methods, random view sampling, and traditional prior active vision approaches, \changes{upgraded to be competitive in our setting.}
\begin{itemize}
  \item \textbf{Single view (neural net)}: has access to only one view, like the starting view provided to the active systems. A feed-forward neural network is used for this baseline, composed from the appropriate components of the \textsc{sensor} and \textsc{classifier} modules of our system. This baseline is entirely pose-agnostic \emph{i.e.}, the same classifier is applied to views from all object poses.
  \item \textbf{Random views (average)}: uses the same architecture as ``Single view (neural net)'', but has access to $T$ views, with successive views being related by randomly selected motions from the same motion set $\mathcal{M}$ available to the active systems.  Its output class likelihood at $t=T$ is the average of its independent estimates of class likelihood for each view.
  \item \textbf{Random views (recurrent)}: uses the same core architecture as our Active RNN method, except for the \textsc{actor} module. In its place, random motions (from $\mathcal{M}$) are selected. Note that this should be a strong baseline, having nearly all aspects of the proposed approach except for the active view selection module.  In particular, it has access to its selected motions in its \textsc{sensor} module, and can also learn to intelligently aggregate evidence over views in its \textsc{aggregator} RNN module.
  \item \textbf{Transinformation}: is closely based on~\cite{Schiele1998-ph}, in which views are selected greedily to reduce the information-theoretic uncertainty of the category hypothesis. We make modifications for our setting, such as using 1024-D CNN features in place of the original receptive field histogram features, and using Monte Carlo sampling to approximate information gain. Each view is classified with pose-specific classifiers. When the class hypothesis is identical between consecutive views, it is emitted as output and view selection terminates. Like most prior approaches, this method relies on a canonical world coordinate space in which all object instances can be registered. Since this is infeasible in the active categorization setting, we treat each instance's coordinates as world coordinates.
  \item \textbf{SeqDP}: is closely based on~\cite{Denzler2002-hf}, and extends~\cite{Schiele1998-ph} using a sequential decision process with Bayesian aggregation of information between views. It runs to a fixed number of views.
  \item \textbf{Transinformation + SeqDP}: combines the strengths of~\cite{Schiele1998-ph} and~\cite{Denzler2002-hf}; it uses Bayesian information aggregation across views, and terminates early when the predicted class remains unchanged at consecutive timesteps.
\end{itemize}

Hyperparameters for all methods were optimized for overall accuracy on a validation set through iterative search over random combinations~\cite{bergstra2012random}.

\subsection{Results}\label{sec:recog_result}

\begin{figure}
  \centering
\begin{tabular}{cc}
  \includegraphics[trim={1.4cm 0.1cm 0.5cm 0.05cm},clip,height=0.20\textwidth]{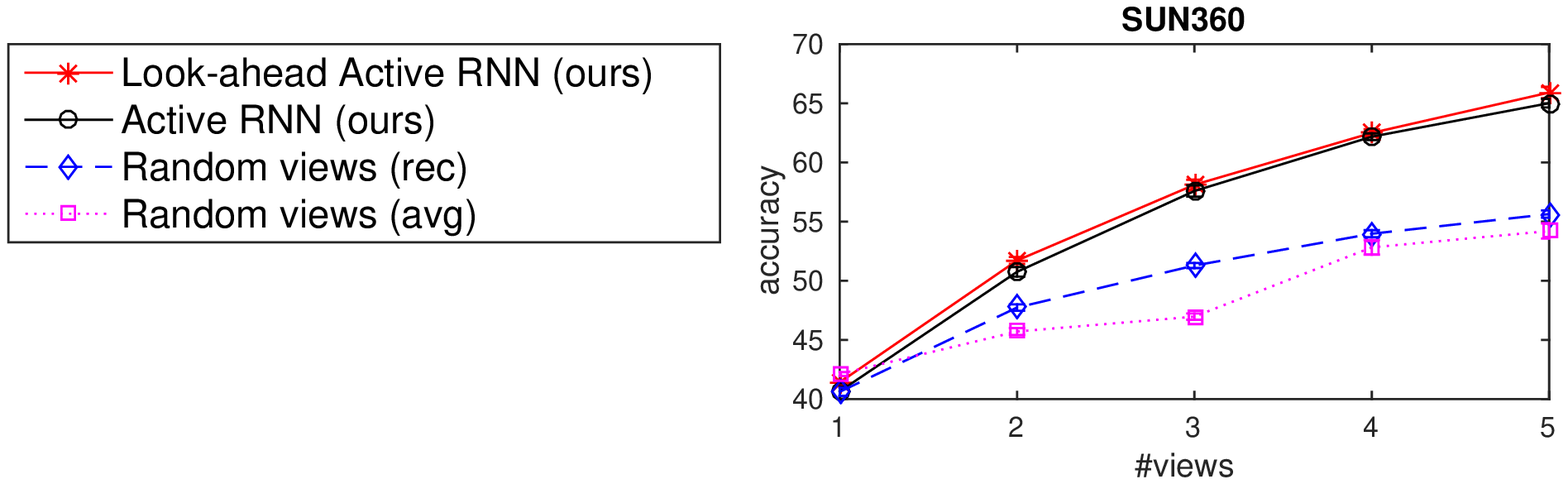}
  \includegraphics[trim={0.5cm 0.1cm 0.5cm 0.05cm},clip,height=0.20\textwidth]{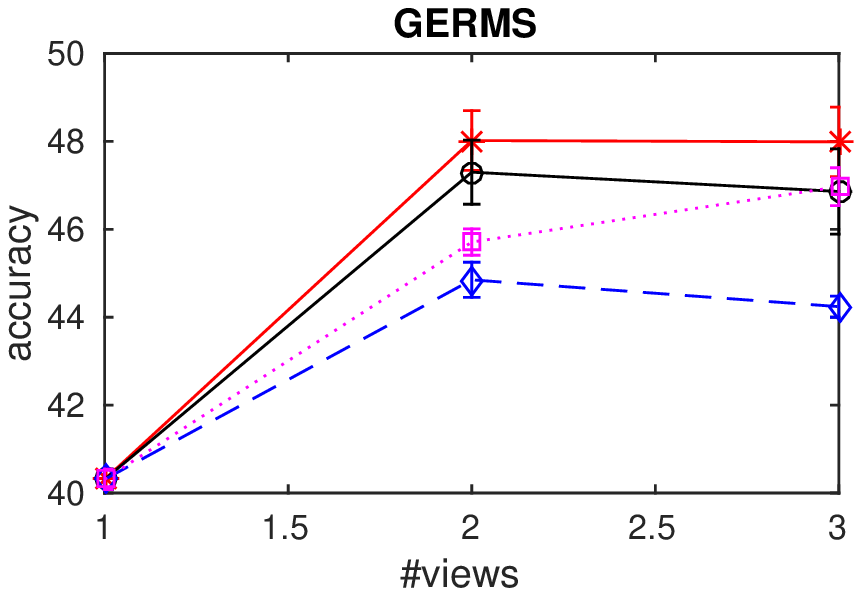}
\end{tabular}
\caption{\small{Evolution of accuracy vs time for various active recognition methods, on SUN360 (left) and GERMS (right)}. \changes{Our methods show steady improvement with additional views, and easily outperform the best baselines. Also see Tab~\ref{tab:recog_result}.}}
  \label{fig:acctime}
\end{figure}

Table~\ref{tab:recog_result} shows the recognition accuracy results for scene categorization (SUN360) and object instance recognition (GERMS), and Figure~\ref{fig:acctime} plots the results as a function of timesteps. Both variants of our method outperform the baselines on both datasets, confirming that our active approach successfully learns intelligent view selection strategies. \changes{Passive baselines, representative of the current standard approaches to visual categorization, perform uniformly poorly, highlighting the advantages of the active setting.} In addition, our Look-ahead active RNN outperforms our Active RNN variant on both datasets, showing the value in simultaneously learning to predict action-conditional next views at the same time we learn the active vision policy. By ``looking before leaping'' our look-ahead module facilitates beneficial knowledge transfer for the active vision task.

On SUN360, even though it represents a much harder active \emph{category recognition} problem, the margins between our method and the random view baselines are pronounced. Furthermore, while the traditional active baselines do show significant improvements from observing multiple views, they fall far short of the performance of our method despite upgrading them in order to be competitive, such as by using CNN features, as described above.

\begin{table}[t]
  \centering
  \resizebox{1\textwidth}{!}{
    \begin{tabular}{ll|ccc|cc}
      \toprule
      \multicolumn{2}{c|}{Method$\downarrow$/Dataset$\rightarrow$} &   \multicolumn{3}{c|}{SUN360}                        & \multicolumn{2} {c}{GERMS}                   \\
      \multicolumn{2}{c|}{Performance measure$\rightarrow$}        & T=2 acc.                & T=3 acc.                & T=5 acc.                & T=2 acc.                & T=3 acc.               \\
      \midrule
      \multirow{2}{*}{\textbf{Passive approaches}} &  Chance                                 & 14.08                   & 14.08                   & 14.08                   & 0.74                    & 0.74                    \\
       &  Single view (neural net)               & 40.12$\pm$0.45          & 40.12$\pm$0.45          & 40.12$\pm$0.45          & 40.31$\pm$0.23          & 40.31$\pm$0.23          \\
      \midrule
      \multirow{2}{*}{\textbf{Random view selection}} &  Random views (average)                 & 45.71$\pm$0.29          & 50.47$\pm$0.37          & 54.21$\pm$0.57          & 45.71$\pm$0.30          & 46.97$\pm$0.43          \\
       &  Random views (recurrent)               & 47.74$\pm$0.27          & 51.29$\pm$0.21          & 55.64$\pm$0.28          & 44.85$\pm$0.40          & 44.24$\pm$0.24          \\
      \midrule
      \multirow{3}{*}{\textbf{Prior active approaches~}} &  Transinformation~\cite{Schiele1998-ph} & 40.69                    & 40.69                   & 44.86                    & 28.83                    & 31.02 \\
      &  SeqDP~\cite{Denzler2002-hf}            & 42.41                  & 42.91                   & 42.08                    & 28.83                    & 28.10 \\
      &  Transinformation + SeqDP               & 44.69                  & 46.91                   & 48.19                    & 29.93                    & 29.56 \\
      \midrule
      \multirow{3}{*}{\textbf{Ours}} &  Active RNN                      & 50.76$\pm$0.41          & 57.52$\pm$0.46          & 65.32$\pm$0.42          & 47.30$\pm$0.73          & 46.86$\pm$0.97          \\
      &  Look-ahead active RNN           & \textbf{51.72$\pm$0.29} & \textbf{58.12$\pm$0.43} & \textbf{66.01$\pm$0.34} & \textbf{48.02$\pm$0.68} & \textbf{47.99$\pm$0.79} \\
      &  Look-ahead active RNN+average   & 49.62$\pm$0.43          & 55.43$\pm$0.38          & 62.61$\pm$0.33               & 47.00$\pm$0.45          & \textbf{48.31$\pm$0.72} \\
      \bottomrule
    \end{tabular}
  }
  \caption{\small{Recognition accuracy for both datasets (neural net-based methods' scores are reported as mean and standard error over 5 runs with different initializations)}}
  \label{tab:recog_result}
\end{table}

On GERMS, our method is once again easily superior to prior active methods. The margins of our gains over random-view baselines are smaller than on SUN360. Upon analysis, it becomes clear that this is due to GERMS being a relatively small dataset.  Not only is (1) the number of active recognition instances small compared to SUN360 (816 vs. 8992), but (2) different views of the same object instance are naturally closer to each other than different views from a SUN360 panorama view-grid (see Fig~\ref{fig:SUN360} and Fig~\ref{fig:GERMS}) so that even single view diversity is low, and (3) there is only a single degree of motion compared to two in SUN360.  As a result, the number of possible reinforcement learning episodes is also much smaller.  Upon inspection, we found that these factors can lead our end-to-end network to overfit to training data (which we countered with more aggressive regularization).  In particular, it is problematic if our method achieves zero training error from just single views, so that the network has no incentive to learn to aggregate information across views well. \KG{Our active results are in line with those presented as a benchmark in the paper introducing the dataset~\cite{GERMS}, and we expect more training data is necessary to move further with end-to-end learning on this challenge.} This lack of data affects our prior active method baselines even more since they rely on \emph{pose-specific} instance classifiers, so that each classifier's training set is very small. This explains their poor performance.

As an interesting upshot, we see further improvements on GERMS by averaging the \textsc{classifier} modules' outputs \ie class likelihoods estimated from the aggregated features at each timestep $t=1,..,T$ (``Look-ahead active RNN + average''). Since the above factors make it difficult to learn the optimal \textsc{aggregator} in an end-to-end system like ours, a second tier of aggregation in the form of averaging over the outputs of our system can yield improvements.
In contrast, since SUN offers much more training data, averaging over per-timestep \textsc{classifier} outputs significantly \emph{reduces} the performance of the system, compared to directly using the last timestep output.  This is exactly as one would hope for a successful end-to-end training. This reasoning is further supported by the fact that ``Random views (average)'' shows slightly poorer performance than ``Random views (recurrent)'' on GERMS, but is much better on SUN360.

\begin{figure}[t]
  \centering
  \includegraphics[width=1\linewidth]{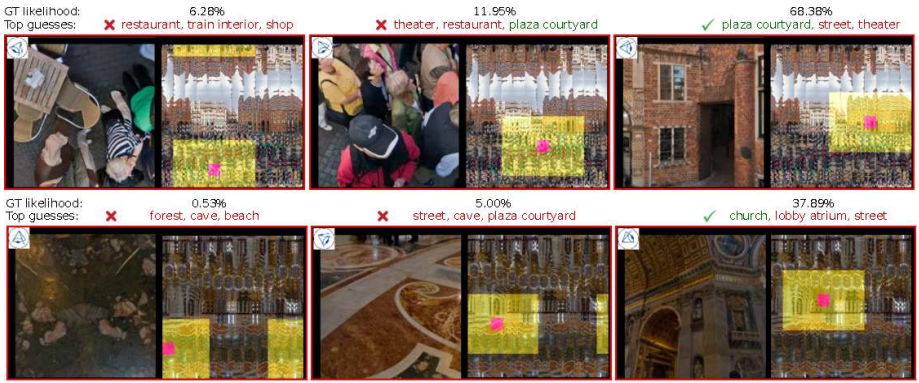}
  \caption{\small{Views selected using our approach on SUN360. Each row, corresponding to a scene, contains 3 red panels corresponding to the selected views at $t=1,2,3$. Each panel shows the current view (left) and position on view grid (pink highlight is current position). In the top row, given the first view, our method makes reasonable but wrong guesses, but corrects itself within 2 moves, by observing the crowd and following their gaze. (More examples in Supp.)}}
  \label{fig:glimpses}
\end{figure}

Indeed, the significant gains of ``Random views (recurrent)'' over ``Random views (average)'' on SUN360 points to an important advantage of treating object/scene categorization as a grounded, sequence-based  decision process.  The ability to intelligently fuse observations over timesteps based on both the views themselves and the camera motions relating them offers substantial rewards. In contrast, the current computer vision literature in visual categorization is largely focused on categorization strategies that process individual images outside the context of any agent motion or sequential data, much like the ``Single view'' or ``Random views (average)'' baselines.  We see our empirical results as an exciting prompt for future work in this space.  They also suggest the need for increased efforts creating large 3-D and video benchmark datasets (in the spirit of SUN360 and GERMS and beyond) to support such vision research, allowing us to systematically study these scenarios outside of robot platforms.

\changes{The result on SUN360 in particular is significant since no prior active recognition approach has been shown to successfully handle any comparably complex dataset. While active categorization is technically challenging compared to instance recognition as discussed in Sec~\ref{sec:related}, datasets like SUN360, containing complex visual data with ambiguous views may actually be most suited to showing the advantages of the active recognition paradigm.} %

\section{Conclusions}

We presented a new end-to-end approach for active visual categorization.
Our framework simultaneously learns (1) how the system should move to improve
its sequence of observations, and (2) how a sequence of future observations
is likely to change conditioned on its possible actions.  We show the impact
on object and scene recognition, where our active approach makes
sizeable strides over single view and passively moving systems.
Furthermore, we establish the positive impact in treating all components of
the active recognition system simultaneously.  All
together, the results are encouraging evidence that modern visual
recognition algorithms can venture further into unconstrained, sequential
data, moving beyond the static image snapshot labeling paradigm.

\small{\noindent\textbf{Acknowledgments:} \changes{This research is supported in part by ONR PECASE N00014-15-1-2291. We also thank Texas Advanced Computing Center for their generous support, and Mohsen Malmir and Jianxiong Xiao for their assistance sharing GERMS and SUN360 data respectively.}}

\bibliographystyle{splncs03}
\bibliography{refs}

\begin{thebibliography}{10}
\providecommand{\url}[1]{\texttt{#1}}
\providecommand{\urlprefix}{URL }

\bibitem{pulkit}
Agrawal, P., Carreira, J., Malik, J.: Learning to see by moving. In: ICCV
  (2015)

\bibitem{aloimonos1988active}
Aloimonos, J., Weiss, I., Bandyopadhyay, A.: Active vision. In: IJCV (1988)

\bibitem{Andreopoulos2009-mh}
Andreopoulos, A., Tsotsos, J.: A theory of active object localization. In: ICCV
  (2009)

\bibitem{Andreopoulos2013-bm}
Andreopoulos, A., Tsotsos, J.: 50 years of object recognition: Directions
  forward. In: CVIU (2013)

\bibitem{Ba2014-rr}
Ba, J., Mnih, V., Kavukcuoglu, K.: Multiple object recognition with visual
  attention. In: ICLR (2015)

\bibitem{bajcsy1988active}
Bajcsy, R.: Active perception. In: Proceedings of the IEEE (1988)

\bibitem{ballard1991animate}
Ballard, D.: Animate vision. In: Artificial Intelligence (1991)

\bibitem{Bazzani2011-cb}
Bazzani, L., Larochelle, H., Murino, V., Ting, J.A., Freitas, N.d.: Learning
  attentional policies for tracking and recognition in video with deep
  networks. In: ICML (2011)

\bibitem{bergstra2012random}
Bergstra, J., Bengio, Y.: Random search for hyper-parameter optimization. In:
  JMLR (2012)

\bibitem{borotschnig1998active}
Borotschnig, H., Paletta, L., Prantl, M., Pinz, A., et~al.: Active object
  recognition in parametric eigenspace. In: BMVC (1998)

\bibitem{Bowling2005-ev}
Bowling, M., Ghodsi, A., Wilkinson, D.: Action respecting embedding. In: ICML
  (2005)

\bibitem{brentano1874psychologie}
Brentano, F.: Psychologie vom empirischen Standpunkte (1874)

\bibitem{butko}
Butko, N., Movellan, J.: Optimal scanning for faster object detection. In: CVPR
  (2009)

\bibitem{callari2001active}
Callari, F., Ferrie, F.: Active object recognition: Looking for differences.
  In: IJCV (2001)

\bibitem{chen2015deepdriving}
Chen, C., Seff, A., Kornhauser, A., Xiao, J.: Deepdriving: Learning affordance
  for direct perception in autonomous driving. In: ICCV (2015)

\bibitem{cohen2014transformation}
Cohen, T.S., Welling, M.: Transformation properties of learned visual
  representations. In: arXiv preprint arXiv:1412.7659 (2014)

\bibitem{Denzler2002-hf}
Denzler, J., Brown, C.M.: Information theoretic sensor data selection for
  active object recognition and state estimation. In: TPAMI (2002)

\bibitem{Dickinson1997-my}
Dickinson, S., Christensen, H., Tsotsos, J., Olofsson, G.: Active object
  recognition integrating attention and viewpoint control. In: CVIU (1997)

\bibitem{Ding2014-co}
Ding, W., Taylor, G.W.: Mental rotation by optimizing transforming distance.
  In: NIPS DL Workshop (2014)

\bibitem{Flynn2015-fm}
Flynn, J., Neulander, I., Philbin, J., Snavely, N.: {DeepStereo}: Learning to
  predict new views from the world's imagery. In: CVPR (2016)

\bibitem{gonzalez2014active}
Garcia, A.G., Vezhnevets, A., Ferrari, V.: An active search strategy for
  efficient object detection. In: CVPR (2015)

\bibitem{helmer2009semantic}
Helmer, S., Meger, D., Viswanathan, P., McCann, S., Dockrey, M., Fazli, P.,
  Southey, T., Muja, M., Joya, M., Little, J., et~al.: Semantic robot vision
  challenge: Current state and future directions. In: IJCAI workshop (2009)

\bibitem{Jayaraman2015-aq}
Jayaraman, D., Grauman, K.: Learning image representations tied to ego-motion.
  In: ICCV (2015)

\bibitem{ssfa}
Jayaraman, D., Grauman, K.: Slow and steady feature analysis: higher order
  temporal coherence in video. In: CVPR (2016)

\bibitem{Kulkarni2015-uh}
Kulkarni, T.D., Whitney, W., Kohli, P., Tenenbaum, J.B.: Deep convolutional
  inverse graphics network. In: NIPS (2015)

\bibitem{Levine2015-gi}
Levine, S., Finn, C., Darrell, T., Abbeel, P.: {End-to-End} training of deep
  visuomotor policies. In: ICRA (2015)

\bibitem{lin2014microsoft}
Lin, T.Y., Maire, M., Belongie, S., Hays, J., Perona, P., Ramanan, D.,
  Doll{\'a}r, P., Zitnick, C.L.: Microsoft {COCO}: Common objects in context.
  In: ECCV (2014)

\bibitem{GERMS}
Malmir, M., Sikka, K., Forster, D., Movellan, J., Cottrell, G.W.: Deep
  {Q-learning} for active recognition of {GERMS}. In: BMVC (2015)

\bibitem{mishra2009active}
Mishra, A., Aloimonos, Y., Fermuller, C.: Active segmentation for robotics. In:
  IROS (2009)

\bibitem{Mnih2014-dg}
Mnih, V., Heess, N., Graves, A., Kavukcuoglu, K.: Recurrent models of visual
  attention. In: NIPS (2014)

\bibitem{paletta2000active}
Paletta, L., Pinz, A.: Active object recognition by view integration and
  reinforcement learning. In: RAS (2000)

\bibitem{ramanathan2011active}
Ramanathan, V., Pinz, A.: Active object categorization on a humanoid robot. In:
  VISAPP (2011)

\bibitem{ranzato2014video}
Ranzato, M., Szlam, A., Bruna, J., Mathieu, M., Collobert, R., Chopra, S.:
  Video (language) modeling: a baseline for generative models of natural
  videos. In: arXiv preprint arXiv:1412.6604 (2014)

\bibitem{ILSVRC15}
Russakovsky, O., Deng, J., Su, H., Krause, J., Satheesh, S., Ma, S., Huang, Z.,
  Karpathy, A., Khosla, A., Bernstein, M., Berg, A.C., Fei-Fei, L.: {ImageNet
  Large Scale Visual Recognition Challenge}. In: IJCV (2015)

\bibitem{Schiele1998-ph}
Schiele, B., Crowley, J.: Transinformation for active object recognition. In:
  ICCV (1998)

\bibitem{sermanet2014attention}
Sermanet, P., Frome, A., Real, E.: Attention for fine-grained categorization.
  In: arXiv (2014)

\bibitem{simonyan2014very}
Simonyan, K., Zisserman, A.: Very deep convolutional networks for large-scale
  image recognition. In: arXiv (2014)

\bibitem{Soatto2009-pk}
Soatto, S.: Actionable information in vision. In: {ICCV} (2009)

\bibitem{Stober2011-sn}
Stober, J., Miikkulainen, R., Kuipers, B.: Learning geometry from sensorimotor
  experience. In: ICDL (2011)

\bibitem{googlenet}
Szegedy, C., Liu, W., Jia, Y., Sermanet, P., Reed, S., Anguelov, D., Erhan, D.,
  Vanhoucke, V., Rabinovich, A.: Going deeper with convolutions. In: CVPR
  (2015)

\bibitem{Vondrick2015-vq}
Vondrick, C., Pirsiavash, H., Torralba, A.: Anticipating the future by watching
  unlabeled video. In: CVPR (2016)

\bibitem{walker2015dense}
Walker, J., Gupta, A., Hebert, M.: Dense optical flow prediction from a static
  image. In: ICCV (2015)

\bibitem{Watter2015-jn}
Watter, M., Springenberg, J.T., Boedecker, J., Riedmiller, M.: Embed to
  control: A locally linear latent dynamics model for control from raw images.
  In: NIPS (2015)

\bibitem{wilkes1992active}
Wilkes, D., Tsotsos, J.: Active object recognition. In: CVPR (1992)

\bibitem{williams1992simple}
Williams, R.: Simple statistical gradient-following algorithms for
  connectionist reinforcement learning. In: JMLR (1992)

\bibitem{wu20153d}
Wu, Z., Song, S., Khosla, A., Yu, F., Zhang, L., Tang, X., Xiao, J.: {3D
  ShapeNets}: A deep representation for volumetric shape modeling. In: CVPR
  (2015)

\bibitem{SUN360}
Xiao, J., Ehinger, K., Oliva, A., Torralba, A., et~al.: Recognizing scene
  viewpoint using panoramic place representation. In: CVPR (2012)

\bibitem{Xu2015-vc}
Xu, K., Ba, J., Kiros, R., Cho, K., Courville, A., Salakhutdinov, R., Zemel,
  R., Bengio, Y.: Show, attend and tell: Neural image caption generation with
  visual attention. In: ICML (2015)

\bibitem{aloim}
Yu, X., Fermuller, C., Teo, C.L., Yang, Y., Aloimonos, Y.: Active scene
  recognition with vision and language. In: CVPR (2011)

\end{thebibliography}

\end{document}